\documentclass[letterpaper, 10 pt, journal, twoside]{ieeetran}

\IEEEoverridecommandlockouts                              

\newif\ifarxiv
\arxivtrue

\usepackage{graphics,graphicx} 
\graphicspath{{Figures/},{./}}

\usepackage{amsmath} 
\usepackage{amssymb}  
\usepackage{xcolor}
\usepackage{algorithm}
\usepackage{algpseudocode}
\usepackage{cite}
\usepackage{dblfloatfix}

\newcommand{\abs}[1]{\left\lvert#1\right\rvert}

\newcommand{\x}{\mathbf{x}}
\newcommand{\xbar}{\Bar{\x}}
\newcommand{\Vx}{\V_{\x}}
\newcommand{\Vxx}{\V_{\x\x}}

\newcommand{\ub}{\mathbf{u}}
\newcommand{\f}{\mathbf{f}}
\newcommand{\fx}{\mathbf{f}_{\mathbf{x}}}
\newcommand{\fu}{\mathbf{f}_{\mathbf{u}}}
\newcommand{\fxx}{\mathbf{f}_{\mathbf{_{xx}}}}
\newcommand{\fuu}{\mathbf{f}_{\mathbf{{uu}}}}
\newcommand{\fux}{\mathbf{f}_{\mathbf{{ux}}}}
\newcommand{\V}{\mathbf{V}}

\newcommand{\inv}[1]{{#1}^{-1}}
\newcommand{\Qx}{\mathbf{Q}_{\mathbf{x}}}
\newcommand{\Qu}{\mathbf{Q}_{\mathbf{u}}}
\newcommand{\Qxx}{\mathbf{Q}_{\mathbf{xx}}}
\newcommand{\Qux}{\mathbf{Q}_{\mathbf{ux}}}
\newcommand{\Quu}{\mathbf{Q}_{\mathbf{uu}}}

\newcommand{\lx}{\mathbf{L}_{\x}}
\newcommand{\lu}{\mathbf{L}_{\ub}}
\newcommand{\lxx}{\mathbf{L}_{\x\x}}
\newcommand{\luu}{\mathbf{L}_{\ub\ub}}
\newcommand{\lux}{\mathbf{L}_{\ub\x}}
\newcommand{\qs}{\mathbf{q}}
\newcommand{\Hb}{\mathbf{H}}
\newcommand{\vb}{\mathbf{v}}
\newcommand{\J}{\mathbf{J}}
\newcommand{\C}{\mathbf{C}}
\newcommand{\taub}{\boldsymbol{\tau}}
\newcommand{\lambdab}{\boldsymbol{\lambda}}

\newcommand{\lambdahb}{\hat{\boldsymbol{\lambda}}}

\newcommand{\Pb}{\mathbf{P}}
\newcommand{\I}{\mathbf{I}}
\newcommand{\Sb}{\mathbf{S}}

\newcommand{\kappab}{\boldsymbol{\kappa}}
\newcommand{\z}{\mathbf{z}}
\newcommand{\X}{\mathbf{X}}
\newcommand{\tb}{\mathbf{t}}
\newcommand{\D}{\mathcal{D}}
\newcommand{\Sc}{\mathcal{S}}

\usepackage{hyperref}
\hypersetup{
    colorlinks=true,
    linkcolor=black,
    filecolor=black,  
    citecolor = blue,
    urlcolor=cyan,
}

\title{
Hybrid Systems Differential Dynamic Programming for Whole-Body Motion Planning of Legged Robots
}

\author{He Li and Patrick M. Wensing 
\thanks{This work was supported by NSF Grant CMMI-1835186.}
\thanks{He Li and Patrick Wensing are with the Department of Aerospace and Mechanical Engineering,
        University of Notre Dame, Notre Dame, IN 46556 USA
        ({\tt\small hli25@nd.edu, pwensing@nd.edu})}%
}

\newcommand{\mynewpage}[1]{}
\begin{document}

\markboth{Author Pre-Print. Please see final RA-L version: \url{https://doi.org/10.1109/LRA.2020.3007475}}
{Li \MakeLowercase{\textit{et al.}}: Hybrid Systems Differential Dynamic Programming}

\maketitle

\begin{abstract}
This paper presents a Differential Dynamic Programming (DDP) framework for trajectory optimization (TO) of hybrid systems with state-based switching. The proposed Hybrid Systems DDP (HS-DDP) approach is considered for application to whole-body motion planning with legged robots. Specifically, HS-DDP incorporates three algorithmic advances: an impact-aware DDP step addressing the impact event in legged locomotion, an Augmented Lagrangian (AL) method dealing with the switching constraint, and a Switching Time Optimization (STO) algorithm that optimizes switching times by leveraging the structure of DDP. Further, a Relaxed Barrier (ReB) method is used to manage inequality constraints and is integrated into HS-DDP for locomotion planning. The performance of the developed algorithms is benchmarked on a simulation model of the MIT Mini Cheetah executing a bounding gait. We demonstrate the effectiveness of AL and ReB for handling switching constraints, friction constraints, and torque limits. By comparing to previous solutions, we show that the STO algorithm achieves 2.3 times more reduction of total switching times, demonstrating the efficiency of our method.

\end{abstract}


\section{Introduction}\label{sec_intro}


\IEEEPARstart{M}{any} tasks in agriculture, construction, defense, and disaster response require mobile robots to traverse irregular terrains and move through narrow passages. The mobility afforded by legged robots makes them exceptionally suitable for these scenarios. 
Practical challenges to unlock their mobility include the highly nonlinear and hybrid nature of their multi-contact dynamics, a need for on-the-fly generation of motion plans, and the management of various constraints.  

Despite these difficulties, many successful algorithms have been developed and tested in simulation and on hardware \cite{bellicoso2018dynamic,wieber2006trajectory,apgar2018fast,wensing2013high,hutter2014quadrupedal,di2018dynamic,kuindersma2016optimization}. Conventional approaches often optimize the Center of Mass (CoM) trajectory and foothold locations using a reduced-order model and adopt QP-based operational space control (OSC) laws \cite{wensing2013high,apgar2018fast,hutter2014quadrupedal} to select joint torques that track the planned trajectories. Widely used reduced-order models include the Linear Inverted Pendulum (LIP) \cite{wieber2006trajectory} and the Spring-Loaded Inverted Pendulum (SLIP) \cite{wensing2013high,apgar2018fast} for determining foothold locations, with the Zero-Moment Point (ZMP) criterion used to enforce admissible CoM trajectories \cite{bellicoso2018dynamic,wieber2006trajectory,apgar2018fast}. Centroidal dynamics models have also been used that consider the linear and angular momentum of the system as a whole \cite{orin2013centroidal,di2018dynamic,kuindersma2016optimization}. Overall, these approaches have the advantage of fast computation, but the complexity of the resulting motions is limited. For example, motions such as standing up from the ground cannot be generated with a LIP model since it neglects all kinematics constraints and assumes constant height and zero angular momentum.

By comparison, whole-body motion planning can generate more complex behaviors. Whereas QP-based OSC only considers the instantaneous effects of joint torques, whole-body motion planning finds a sequence of torques by solving a finite-horizon trajectory optimization (TO) problem, potentially enabling recovery from larger disturbances. Despite the appeal of this approach, the curse of dimensionality caused by the high-dimensional state space of legged robots has prevented it from being popular. Recent results (e.g., \cite{tassa2012synthesis}) using Differential Dynamic Programming (DDP) \cite{mayne1966second} have shown great promise for online use. Many complementary DDP advances have been proposed, demonstrating robustness for disturbance rejection \cite{morimoto2003minimax} and real-time performance for whole-body motion planning \cite{budhiraja2018differential,farshidian2017efficient,koenemann2015whole,neunert2018whole}.

\begin{figure}
    \centering
    \includegraphics[width=0.7\columnwidth]{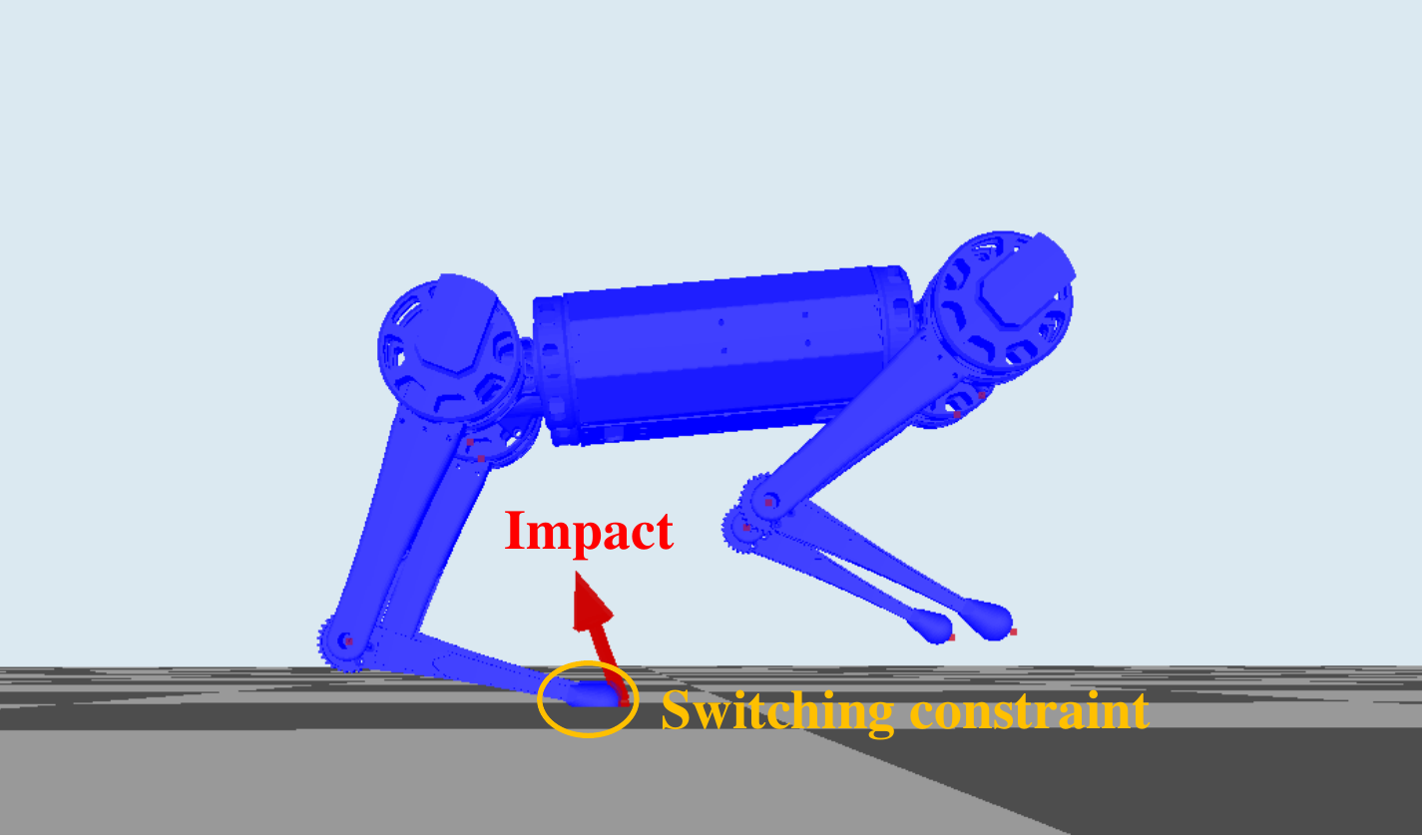}
    \vspace{-10px}
    \caption{Mini Cheetah bounding. This paper develops coordinated advances to the Differential Dynamic Programming (DDP) algorithm for trajectory optimization in hybrid systems. In particular, the methods focus on handling the impact event, the associated switching constraints, and the inequality constraints such as torque limits and friction constraints.
    }
    \label{fig_model}
    \vspace{-12px}
\end{figure}

Unlike conventional direct methods, which optimize over all decision variables together, DDP adopts a \textit{divide and conquer} strategy by successively solving much smaller optimization problems \cite{mayne1966second}. This feature makes DDP exceptionally suitable for problems with long time horizons because the computational effort scales linearly with time as opposed to quadratic or cubic growth with many nonlinear programming (NLP) approaches to TO.
Since DDP is a shooting method, the algorithm can also be terminated at any time while still giving a physically valid trajectory. These features and the successes of \cite{farshidian2017efficient,koenemann2015whole,neunert2018whole} together suggest the promise of DDP for online MPC over other direct methods.

Despite these benefits and promise, there are some difficulties for DDP to be used in legged locomotion planning, such as dealing with the impact discontinuity and managing various constraints. The first difficulty is addressed in \cite{tassa2012synthesis} by approximating the impact discontinuity with a smooth transition, and in \cite{farshidian2017efficient} by ignoring the impact in DDP but compensating for this simplification with a feedback controller. These approaches either do not have experimental evidence or present a robustness issue. Other previous work has contributed to attacking the second difficulty by leveraging constraint-handling techniques from NLP. Box input constraints are handled by solving QPs with a Projected Newton algorithm in \cite{tassa2014control}. A penalty method is used in \cite{farshidian2017efficient} to satisfy state constraints. This method, however, has a numerical ill-conditioning problem that results when penalty coefficients are large. Augmented Lagrangian (AL) methods (e.g., \cite{Howell19}) resolve this issue by adding a linear multiplier term. Lantoine et al.~\cite{lantoine2012hybrid} proposed a DDP algorithm that handles terminal state constraints using AL, motivating their use to address the state-based switching for hybrid systems in this work.

In this paper, we propose a Hybrid Systems DDP (HS-DDP) approach that extends the applicability of DDP to hybrid systems. In particular, HS-DDP includes three algorithmic advances: an impact-aware DDP step that addresses impact discontinuities, an AL method for switching constraints, and a switching time optimization (STO) strategy. Further, in order to deal with the many inequality constraints in legged locomotion, a relaxed barrier (ReB) method \cite{hauser2006barrier,grandia2019feedback} is adopted and is integrated within HS-DDP. The developed algorithms are benchmarked in simulation on Mini Cheetah bounding, as shown in Fig.~\ref{fig_model}. The developed algorithms are extendable to general gaits such as trotting and galloping etc., and to other platforms such as bipeds and manipulators.

The structure of this paper is as follows. DDP background and the hybrid dynamics formulation are given in Sections~\ref{sec_back_DDP} and \ref{sec_back_model}. Section~\ref{sec_theory} discusses the main contributions of this letter, which extend DDP to hybrid systems. Section~\ref{sec_sim} analyzes the performance of the proposed algorithm in terms of constraint handling and efficiency of the STO as applied to quadruped bounding. Section \ref{sec_conclusion} provides a closing discussion.

\mynewpage

\section{Background: Differential Dynamic Programming}\label{sec_back_DDP}
This section gives a brief introduction to DDP following \cite{tassa2012synthesis}. Readers are referred to \cite{mayne1966second} for detailed derivation. 
The goal of DDP is to find an optimal control sequence $\mathbf{U}^*=\{\ub^*_k\}_{k=0}^{N-1}$ that minimizes a cost function $J$ of the form
\begin{equation}\label{eq_costfun}
    J(\mathbf{U}) = \sum_{k=0}^{N - 1} {L(\x_k,\ub_k)} + \Phi(\x_N)
\end{equation}
where $\{\x_k\}_{k=0}^N$ denotes the state trajectory, $L$ denotes the running cost, and $\Phi$ denotes the terminal cost. The trajectory $\{\x_k\}_{k=0}^N$ is subject to the discretized dynamics
\begin{equation}\label{eq_dyn}
    \x_{k+1} = \mathbf{f}(\x_k,\ub_k)
\end{equation}
where $\x$ and $\ub$ respectively denote the state and control variables. DDP recursively finds $\mathbf{U}^*$ by repeatedly executing a forward sweep and a backward sweep. Given a nominal control sequence, the forward sweep computes a nominal trajectory and the associated dynamics derivatives. A backward sweep is then executed to generate a policy that is used to update the control sequence. As this process continues, the control sequence (locally) converges to $\mathbf{U}^*$. Since DDP optimizes only over the control sequence, it can be classified as a direct shooting method. Interested readers may refer to \cite{diehl2006fast} for a discussion of tradeoffs with other direct methods.

Denote $V(\x_k)$ the value function (i.e., optimal cost-to-go) at time step $k$. Using Bellman's principle of optimality, $V(\x_k)$ is given recursively backward in time:
\begin{equation}\label{eq_Bellman}
    V(\x_{k}) = \min_{\ub_k}[ \underbrace{L(\x_k, \ub_k) + V(\x_{k+1})}_{Q(\x_k, \ub_k)}]\,.
\end{equation}
Attempts to solve (\ref{eq_Bellman}) directly are difficult since an analytical expression for $V(\x_k)$ is rarely possible due to nonlinearity of $\f(\x_k,\ub_k)$. To avoid this problem, DDP considers the variation of $Q(\x_k,\ub_k)$ around a nominal state-control pair $(\hat{\x},\hat{\ub})$ under the perturbation $(\delta\x, \delta\ub)$. The resulting variation $\delta Q(\delta\x,\delta\ub)$ is approximated to the second order as:
\newcommand{\pmwspace}{.5ex}
\begin{multline}\label{eq_Qapprox}
    \delta Q(\delta\x, \delta\ub) \approx
    \frac{1}{2}
    \begin{bmatrix}
    1 \\[\pmwspace] \delta\x \\[\pmwspace] \delta\ub
    \end{bmatrix}^T
    \begin{bmatrix}
        0   &   \Qx^T   &   \Qu^T \\[\pmwspace]
        \Qx &   \Qxx    &   \Qux^T \\[\pmwspace]
        \Qu &   \Qux    &   \Quu    \end{bmatrix}
     \begin{bmatrix}
    1 \\[\pmwspace] \delta\x \\[\pmwspace] \delta\ub
    \end{bmatrix},
\end{multline}
where 
\vspace{-6px}
\begin{subequations}\label{eq_Qs}
\begin{align}
    \Qx &= \lx + \fx^T\Vx',\\
    \Qu &= \lu + \fu^T\Vx',\\
    \Qxx &= \lxx + \fx^T\Vxx'\fx + \Vx' \cdot  \fxx,\\
    \Quu &= \luu + \fu^T\Vxx'\fu + \Vx' \cdot  \fuu,\\
    \Qux &= \lux + \fu^T\Vxx'\fx + \Vx' \cdot \fux,
\end{align}
\end{subequations}
in which the subscripts indicate the partial derivatives and the prime indicates the next time step. Note that $\fxx$, $\fuu$ and $\fux$ generally are tensors. The notation `$\cdot$' denotes matrix-tensor multiplication. Omitting the third terms in the last three equations gives rise to the iLQR algorithm, which enables faster iterations but loses quadratic convergence properties. We employ iLQR in this work and use the algorithm proposed in \cite{jain1993linearization,carpentier2018analytical} to efficiently compute $\f_{\x}$ and $\fu$.

Optimizing $\delta Q(\delta\x, \delta\ub)$ over $\delta\ub$ results in the optimal control increment $\delta\ub^*$ around the nominal control $\hat{\ub}$ as
\begin{equation}\label{eq_optdu}
     \delta\ub^* = -\Quu^{-1}(\Qu + \Qux\delta\x) = \kappab + \mathbf{K}\delta\x,
\end{equation}
where $\kappab$ is the step direction and $\mathbf{K}$ is the feedback gain. Substituting (\ref{eq_optdu}) into the equation (\ref{eq_Qapprox}) results in update equations for the local quadratic model of $V$ according to
\begin{subequations}\label{eq_updateV}
\begin{align}
    \Delta V &= \Delta V'  + \frac{1}{2}\Qu^T\Quu^{-1}\Qu, \label{eq_deltaV}\\
    \Vx &= \Qx - \Qux^T\Quu^{-1}\Qu, \label{eq_Vx}\\
    \Vxx &= \Qxx - \Qux^T\Quu^{-1}\Qux \label{eq_Vxx}.
\end{align}
\end{subequations}
where $\Delta V$ denotes the expected cost reduction. 

The equations (\ref{eq_Qs}) and (\ref{eq_updateV}) are computed recursively starting at the final state, constituting the backward pass of DDP. The nominal control is then updated using the resulting control policy (\ref{eq_optdu}) as follows,
\begin{equation}\label{eq_uff}
    \ub_k = \hat{\ub}_k + \epsilon\kappab + \mathbf{K}(\x_k - \hat{\x}_k),
\end{equation}
where $0<\epsilon\leq 1$ is a line search parameter, $(\hat{\x},\hat{\ub})$ and $(\x,\ub)$ respectively are the nominal and new state-control pair. A backtracking line search method is used to select $\epsilon$ \cite{tassa2012synthesis} and a regularization strategy as in \cite{tassa2012synthesis} is employed, ensuring a decrease of the cost in each iteration. The forward-backward process above is repeated until the algorithm converges or a certain number of iterations is reached.

\section{Background: Dynamics Modeling}
\label{sec_back_model}
\begin{figure}[!t]
\centering
\includegraphics[width=.85\columnwidth]{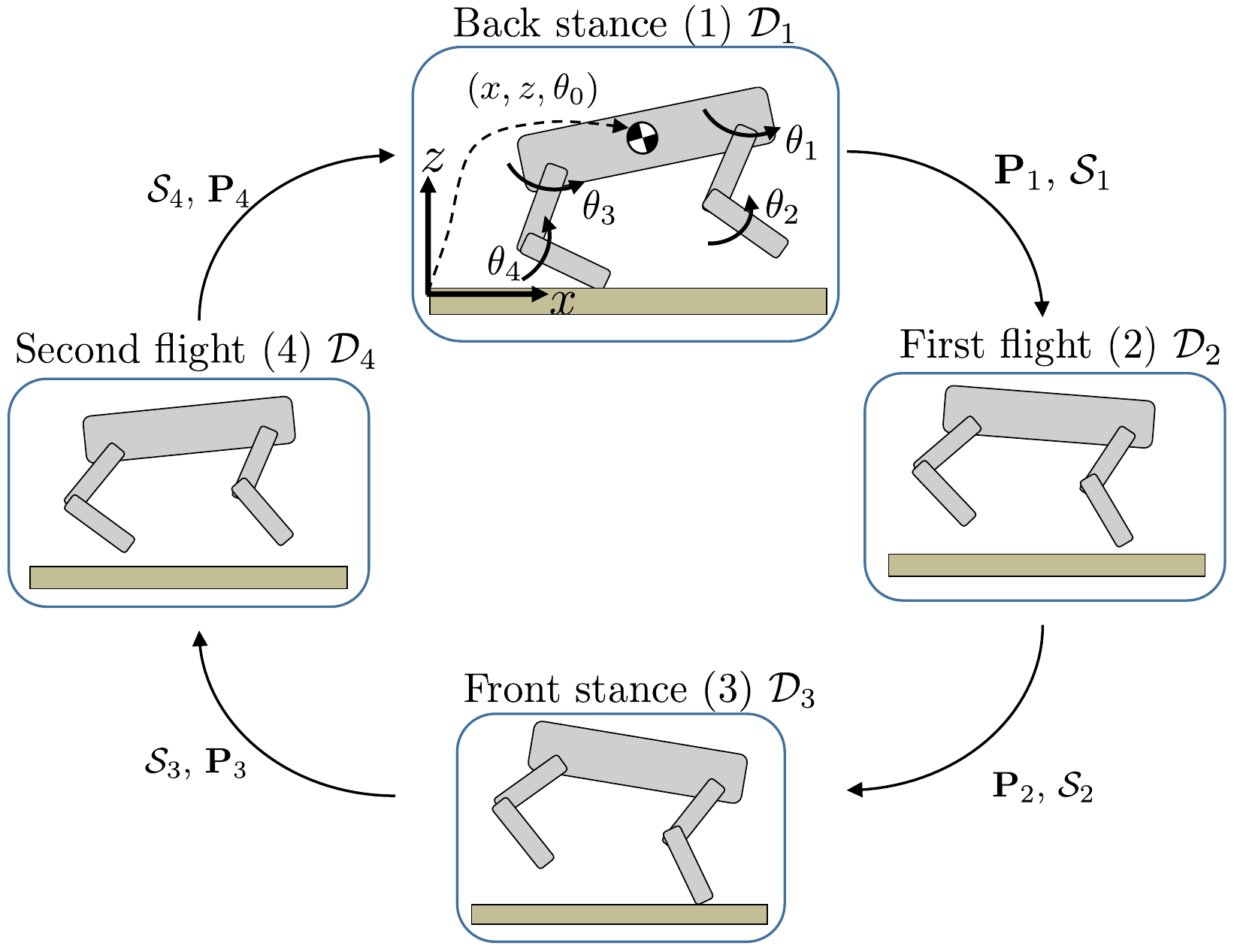}
\vspace{-8px}
\caption{Mode sequence of a quadruped bounding gait. The gait cycle is assumed to start from the back stance for simplicity of presentation. The generalized coordinates for this 2D quadruped are $\qs = [x,z,\theta_0,\theta_1,\theta_2,\theta_3,\theta_4]^T$.}
\label{fig_gait_cycle}
\vspace{-5px}
\end{figure}

This section presents a hybrid system model for bounding quadrupeds. Figure \ref{fig_gait_cycle} shows one gait cycle of quadruped bounding with four continuous modes and a reset map between every two consecutive modes. Denote $\mathcal{P}(n)=\{1,\cdots,n\}$ the mode sequence where $n$ represents the total number of modes. Then, $\mathcal{P}(4)$ denotes one gait cycle. The continuous dynamics in mode $i$, denoted by $\Bar{\f}_i$, takes place on domain $\mathcal{D}_i$. The reset map $\Pb_{i}$ takes place on the switching surface $\Sc_i$ at the boundary of $\mathcal{D}_i$. Mathematical definitions of $\mathcal{D}_i$ and $\Sc_i$ are introduced later. Denote $\qs$ the generalized coordinates of the quadruped and $\x = [\qs^T, \vb^T]^T$ the state vector where $\vb = \dot{\qs}$. The hybrid model is given as
\begin{equation}\label{eq_hybridDyn}
    \begin{cases}
    \dot{\x} &= \bar{\f}_i(\x,\ub),\ \x\in\D_i \\
    \x^+ &= \Pb_{i}(\x^-), \ \x^- \in \Sc_i
    \end{cases},
\end{equation}
where `-' and '+' indicate pre- and post-transition states.  

Denote $c_i$ the contact foot in mode $i$ and $\Bar{c}_i$ the other foot. During a flight mode, $c_i$ represents the foot scheduled to touch down at the end of flight. The sets $\mathcal{D}_i$ and $\Sc_i$ are defined for one gait cycle as in Fig.~\ref{fig_gait_cycle},
\begin{subequations}\label{eq_domain}
\begin{align}
    \D_i &= \{\x\in\mathcal{TQ} \, | \, g_{c_i}(\x)=0,  \dot{g}_{c_i}(\x)=0\}, i = 1, 3,\\
    \D_i &= \{\x\in\mathcal{TQ} \, | \, g_{c_i}(\x)>0,  g_{\Bar{c}_i}(\x)>0 \}, i = 2, 4, \\
    \Sc_i &= \{\x\in\mathcal{TQ} \, | \, g_{c_i}(\x)=0, \abs{\dot{g}_{c_i}(\x)} \neq 0 \}, \forall i,
\end{align}
\end{subequations}
where $g(\cdot)$ is a function measuring the vertical distance of the corresponding foot to the ground, $\mathcal{TQ}$ denotes the tangent bundle of the configuration space $\mathcal{Q}$. 
Although the hybrid model (\ref{eq_hybridDyn}) and (\ref{eq_domain}) considers one gait cycle for simplicity, it can be extended to multiple gait cycles.

\subsection{Continuous Dynamics}
The continuous dynamics in \eqref{eq_hybridDyn} varies depending on which legs are in stance. However, these dynamics can be formulated with a unified structure as follows:
\begin{equation}\label{eq_phaseDyn}
        \begin{bmatrix}
                \Hb & -\J_{c_i}^T \\
                -\J_{c_i} & \mathbf{0}
        \end{bmatrix}
            \begin{bmatrix}
                \Ddot{\qs} \\
                \lambdab_{c_i}
            \end{bmatrix}
            =   \begin{bmatrix}
                    \Sb^T\taub - \mathbf{C}\Dot{\qs} - \taub_g\\
                    \Dot{\J}_{c_i}\Dot{\qs}
                \end{bmatrix},
    \end{equation}
where $\Hb$, $\C \Dot{\qs}$, $\taub_g$, $\Sb$, and $\taub$ denote the inertia matrix, Coriolis force, gravity force, selection matrix, and actuation torque, respectively. $\J_{c_i}$ and $\lambdab_{c_i}$ represent the contact Jacobian and contact force associated with the contact foot $c_i$. The matrix on the left side of (\ref{eq_phaseDyn}) is known as the KKT matrix, since the equation (\ref{eq_phaseDyn}) can be obtained via KKT conditions \cite{budhiraja2018differential}. When the robot is in flight, $\J_{c_i}$ and $\lambdab_{c_i}$ are not meaningful anymore, and the KKT matrix degenerates to the inertia matrix $\Hb$. The state-space representation of (\ref{eq_phaseDyn})
is obtained by pre-multiplying both sides of (\ref{eq_phaseDyn}) by the inverse of the KKT matrix 
and separating out the solution for $\ddot{\qs}$.

\subsection{Reset Maps}
While the generalized coordinates remain unchanged across impact events, velocities change instantaneously at each \textit{touch down}. The impact dynamics are modeled as
\begin{equation}\label{eq_impctDyn}
    \begin{bmatrix}
        \Hb & -\J_{c_{i}}^T\\
        -\J_{c_{i}} & \mathbf{0}
    \end{bmatrix} 
    \begin{bmatrix}
    \vb^+\\
    \lambdahb_{c_{i}}
    \end{bmatrix} = 
    \begin{bmatrix}
    \Hb\vb^-\\
    e\J_{c_{i}}\vb^-
    \end{bmatrix},
\end{equation}
where $e\in [0, 1]$ denotes the coefficient of restitution. Perfect inelastic collision with  $e=0$ is assumed in this work, meaning that the contact foot sticks to the ground after impact. The vector $\lambdahb_{c_{i}}$ denotes the impulse acting on the contact foot that is scheduled to touch down at the end of flight. Note that there is no control present in the model (\ref{eq_impctDyn}) since the actuators cannot generate impulsive outputs. By separating $\vb^+$ out, the state-space representation of the reset map at impact is $\x^+ = \Pb_{i}(\x^-)$ where 
\begin{equation}\label{eq_Xproj}
    \Pb_{i}(\x^{-}) = 
    \begin{bmatrix}
        \I & \mathbf{0}\\
        \mathbf{0} & \I-\inv{\Hb}\J_{c_{i}}^T(\J_{c_{i}}\inv{\Hb}\J_{c_{i}}^T)^{-1}\J_{c_{i}}
    \end{bmatrix}
    \mathbf{x}^-.
\end{equation}
Note that the transition from stance to flight is continuous, and, thus, $\Pb_{i}(\x^-) = \x^-$ when $i$ denotes a stance phase.

\subsection{Time-Switched Hybrid System}
We associate the pre-determined mode sequence $\mathcal{P}(n)$ with a switching time vector $\tb = [t_1, \cdots, t_n]$ where $t_i$ represents the terminating time of the $i^{\text{th}}$ mode. Along any trajectory of the state-switched hybrid system,  (\ref{eq_hybridDyn}) and (\ref{eq_domain}) are equivalent to the time-switched hybrid system:
\begin{equation}\label{eq_switchDyn}
    \begin{cases}
    \dot{\x}(t) &= \Bar{\f}_i(\x(t),\ub(t)),\ t\in[t_{i-1}^+, t_{i}^-] \\
    \x^+(t) &= \Pb_{i}(\x^-(t)) , \ t\in[t_{i}^-, t_{i}^+]\\
    \end{cases},
\end{equation}
with the enforced switching constraint:
\begin{equation}\label{eq_constr}
    g_{c_i}(\x(t^-_{i})) = 0.
\end{equation}
In this work, this time-switched reformulation is considered, where variables $t_i$ are optimized under switching constraints. 

\section{Hybrid Systems Differential Dynamic Programming}\label{sec_theory}
\begin{figure}[t]
\center
\includegraphics[width=\columnwidth]{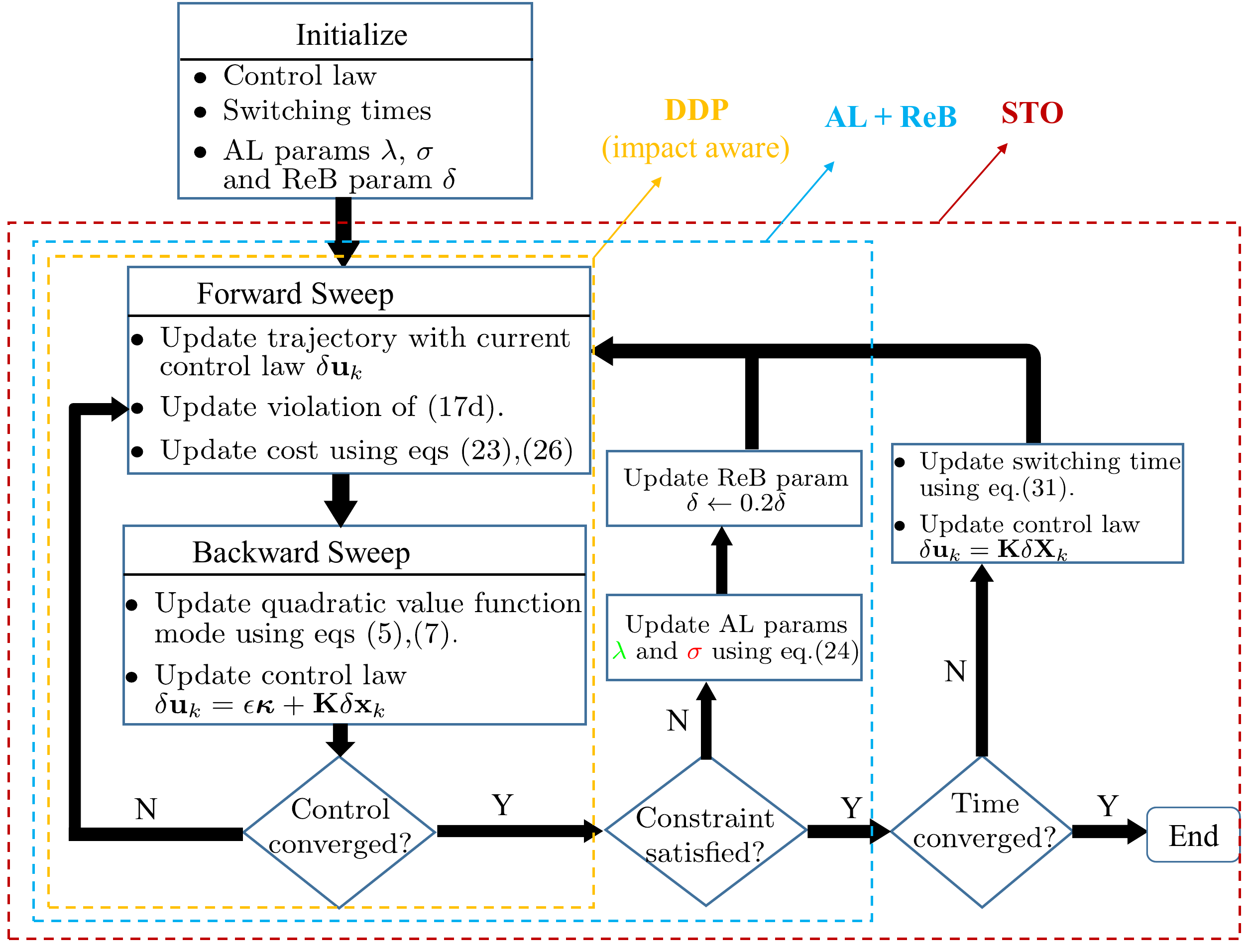}
\vspace{-8px}
\caption{Overview of the HS-DDP algorithmic framework.}
\label{fig_HSDDP}
\vspace{-5px}
\end{figure}

This section discusses three algorithmic advances for HS-DDP and presents the ReB method for inequality constraints. An overview of HS-DDP and ReB is shown in Fig.~\ref{fig_HSDDP}. The HS-DDP algorithm takes a two-level optimization strategy. In the bottom level, the switching times are fixed and the AL algorithm is executed. This algorithm continuously calls the impact-aware DDP. Once DDP converges, the constraint violations are remeasured and added to the cost function, and another DDP call is executed. The AL algorithm terminates when all switching constraints are satisfied. The output from this loop is then utilized by the STO algorithm to update the switching times. This process repeats until the switching times are optimal. The ReB algorithm is executed whenever the AL algorithm is executed. The entire framework is presented to plan trajectories for the quadruped bounding model  introduced in the previous section.

\subsection{Whole-body Motion Planning Problem}
\label{subsec:WholeBodyPlanning}
To find an optimal trajectory, we formulate a TO problem
\begin{equation}\label{eq_optswitch}
\min_{\ub(\cdot),\tb}  \sum_{i=1}^{n} \left[ \int_{t^+_{i-1}}^{t^-_i} l_i(\x(t), \ub(t) ) {\rm d}t + \Phi_i(\x(t^-_i)) \right]
\end{equation}
where $l_i$ and $\Phi_i$ respectively denote the continuous-time running cost 
and the terminal cost for the $i^{\text{th}}$ mode. In whole-body TO, the minimization of (\ref{eq_optswitch}) is subject to the full-order dynamics (\ref{eq_switchDyn}) and other various constraints. A common way to solve (\ref{eq_optswitch}) is to formulate a discrete-time optimal control problem (OCP) with integration time step $h$ as follows
\begin{subequations}\label{eq_switchoptDT}
\begin{align}
    \min_{\mathbf{U},\tb} \ \ \ \ &J(\mathbf{U},\tb) \\
    \text{subject~to} \ \ \ \ &\x_{k+1} = \f_i(\x_k,\ub_k),\label{subeq_dyn}\\ 
    &\x_{N_{i}}^+ = \Pb_{i}(\x_{N_{i}}^-),\label{subeq_reset}\\
    &g_{c_i}(\x_{N_{i}}^-) = 0, \label{subeq_switch}\\
    &\abs{u_{k,j}}\leq u_{\rm{max}},\label{subeq_torque_limit}\\
    &\lambda_{c, z}\geq 0,\\
    &\abs{\lambda_{c,x}}\leq \mu\lambda_{c,z}, \label{subeq_friction}
\end{align}
\end{subequations}
where 
\begin{equation}
    J(\mathbf{U},\tb) = \sum_{i=1}^n \Big(\Phi_i(\x_{N_i}^-) + \sum_{k=N_{i-1}}^{N_i - 1} L_i(\x_k, \ub_k) \Big),
\end{equation}
and $L_i = hl_i$ approximates the integral of the running cost over integration time step $h$, and $N_i=\frac{t_i}{h}$ denotes the number of time steps in the time horizon up to the $i^{\text{th}}$ mode. Equations (\ref{subeq_dyn}) and (\ref{subeq_reset}) represent the dynamics and reset map constraints, respectively, where $\f_i$ is obtained via forward integration of $\bar{\f}_i$. This work uses a forward Euler method, but all algorithmic advances hold with other integration schemes. Equation (\ref{subeq_switch}) represents the switching constraint, and inequalities (\ref{subeq_torque_limit})-(\ref{subeq_friction}) represent the torque limit, non-negative normal GRF, and friction cone constraint, respectively.

\subsection{Impact-Aware Value Function Update in DDP}\label{sec_HSDDP}
Impact-aware DDP extends DDP to address the impact effect, but does not consider constraints (\ref{subeq_switch}) - (\ref{subeq_friction}). While the impact-aware DDP executes the same forward sweep as DDP, it modifies the update equations (\ref{eq_updateV}) for the quadratic value function model at the switching surface. Suppose that $\Delta V$, $\Vx$, and $\Vxx$ are known at $t_i^+$, which can be computed from DDP. The dependency of all variables on $i$ is ignored here for simplicity. Since there is no control applied at $t_i^-$, according to Bellman's Principle of Optimality
\begin{equation}\label{eq_Bellman_trans}
    V(\x^-) = \Phi(\x^-) + V(\x^+).
\end{equation}
Since $\x^-$ and $\x^+$ can be computed from the forward sweep, the variation of (\ref{eq_Bellman_trans}) around $\x^-$ and $\x^+$ is considered, i.e.,
\begin{equation}\label{eq_Bellmanvar_trans}
    V(\x^-+\delta \x^-) = \Phi(\x^- +\delta \x^-) + V(\x^+ +\delta \x^+),
\end{equation}
where
\begin{equation}\label{eq_deltax+}
    \delta \x^+ = \Pb(\x^-+\delta \x^-) - \Pb(\x^-)\approx \Pb_{\x}\delta\x^-
\end{equation}
in which $\Pb_{\x}$ is the Jacobian of $\Pb$ with respect to $\x^-$. Approximating both sides of (\ref{eq_Bellmanvar_trans}) to the second order, we obtain
\vspace{-10px}
\begin{subequations}\label{eq_update_HDDP}
\begin{align}
\Delta V^- &= \Delta V^+,\\
    \V_{\mathbf{xx}}^- & \approx \Phi_{\x\x^-} + \Pb_{\x}^T\V_{\x\x}^+ \Pb_{\x}, \\
     \V_{\x}^- &= \Phi_{\x^-} + \Pb_{\x}^T\V_{\x}^+.
\end{align}
\end{subequations}
The equations (\ref{eq_update_HDDP}) establish the model update equations at the switching surface, which, together with (\ref{eq_updateV}), constitute the model update equations of $V$ for hybrid systems. 

\subsection{Augmented Lagrangian for Switching Constraints}\label{sec_AL}
The impact-aware DDP solves unconstrained optimization problems. Nevertheless, it can be combined with various constraint-handling techniques from NLP for constrained optimization. In this section, we are particularly interested in the switching equality constraint (\ref{subeq_switch}). Penalty methods \cite{farshidian2017efficient} to manage this constraint add a squared term of the constraint violation to the cost function. However, a numerical ill-conditioning issue could happen as the penalty increases. An AL method is employed in this work, which, in addition to the quadratic term, adds a linear Lagrange multiplier term to the cost function, avoiding the numerical ill-conditioning.  

With the AL technique, the cost function now becomes
\begin{equation}\label{eq_AL_cost}
    J(\mathbf{U},\tb) + \left(\frac{\sigma_{\eta}}{2}\right)^2 \sum_{i\in \mathcal{I}_c}g^2_{c_i}(\x^-_{N_i}) + \sum_{i\in \mathcal{I}_c}\lambda_{\eta,i}g_{c_{i}}(\x^-_{N_i})
\end{equation}
where $\mathcal{I}_c$ denotes the set of all touch down indices, $\sigma$ and $\lambda$ denote the penalty and the Lagrange multipliers, respectively. The subscript `$\eta$' is the AL iteration. The AL approach begins with certain initial values for $\sigma$ and $\lambda$, and solves the resulting TO problem using impact-aware DDP. The parameters $\sigma$ and $\lambda$ are then updated and the new TO problem is re-solved using the previous optimal control as a warm start. The update equations are
\begin{equation}
    \sigma_{\eta+1} = \beta\sigma_{\eta} ~~{\rm and}~~ 
    \lambda_{\eta+1,i} = \lambda_{\eta,i} + \sigma_{\eta}g_{c_i}(\x^-_{N_i}), \label{eq_updateLM}
\end{equation}
where $\beta>1$ is the penalty update parameter. This process is repeated until $g_{c_i}$ is within the threshold $\epsilon_{AL}$. To make a distinction, one execution of the forward sweep and backward sweep of DDP is called one DDP iteration. Pseudocode for the AL algorithm is shown in Algorithm \ref{alg_AL}.

\begin{algorithm}[tb]
    \caption{Pseudocode combining AL and ReB}
    \label{alg_AL}
    \begin{algorithmic}[1] 
        \State \textbf{Given}
        \State Mode sequence $\mathcal{P}(n)$ and switching time $\tb$.
        \State Cost function, dynamics, and switching constraints in (\ref{eq_switchoptDT}).
        \State Initial control sequence $\mathbf{U}$ (e.g., zeros).
        \State \textbf{Initialization}
        \State AL parameters $\sigma$, $\lambda_i$, $\beta$, and ReB parameters $\epsilon_B$, $\delta$
        \State Run DDP to convergence and compute $g_{c_i}$.
        \While {$\sum g_{c_i}^2(\x^-_{N_i})>\epsilon_{AL}$}
        \State Update $\sigma \leftarrow \beta\sigma$, $\lambda\leftarrow \lambda + \sigma g_{c_i}$, $\delta\leftarrow 0.2\delta$.
        \State Update (\ref{eq_AL_cost}) and (\ref{eq_opt_barrier}).
        \State Update initial guess for DDP.
        \State Run DDP to convergence.
        \State Compute $g_{c_i}(\x^-_{N_i})$.
        \EndWhile
    \end{algorithmic}
\end{algorithm}

\subsection{Relaxed Barrier Function for Inequality Constraints}
\label{subsec_relaxed_barrier}

We employ a relaxed barrier (ReB) method \cite{hauser2006barrier, grandia2019feedback} to manage the inequality constraints in (\ref{eq_switchoptDT}). Given any inequality constrained optimization problem as below
\begin{equation}\label{eq_constr_opt}
    \begin{aligned}
        \min \ \ \ \ &f(\x) \\
    \text{subject~to} \ \ \ \ &c_j(\x)\geq 0, j=1,\cdots,m,
    \end{aligned}
\end{equation}
ReB attacks (\ref{eq_constr_opt}) by successively solving the unconstrained optimization
\begin{equation}\label{eq_opt_barrier}
    \begin{aligned}
        \min_{\x} \ \ \ \ &f(\x) + \epsilon_B\sum_{j=1}^m B_{\delta}(c_j(\x)),
    \end{aligned}
\end{equation}
where $\epsilon_B >0$ is a weighting parameter and
\begin{equation}\label{eq_relaxed_barrier}
    B_{\delta}(z) = 
        \begin{cases}
        -\log(z) & z>\delta \\
        \frac{k-1}{k}\Big[ \big(\frac{z-k\delta}{(k-1)\delta}  \big)^k-1 \Big]-\log\delta & z\leq \delta
        \end{cases},
\end{equation}
is called a ReB function where $\delta > 0$ is the relaxation parameter. The function $B_{\delta}(z)$ smoothly extends the logarithmic barrier function $-\log(z)$ over the entire real domain with a polynomial of order $k$. In many cases, $k=2$ works well \cite{hauser2006barrier}. Consequently, when applied to a TO problem, the ReB method allows the objective function to be evaluated for an infeasible trajectory, which cannot be done with a standard barrier method. Note that $\delta$ is updated toward zero in an outer loop. This drives the resulting trajectory toward feasibility.

With this technique, the inequality constraints (\ref{subeq_torque_limit}) - (\ref{subeq_friction}) are turned into ReB functions and added to the objective function $J(\mathbf{U}, \tb)$. Combing this technique with AL, the constrained TO problem (\ref{eq_switchoptDT}) is converted into an unconstrained optimization problem, which is solved using the impact-aware DDP. The AL parameters $\lambda$, $\sigma$ and the ReB parameter $\delta$ are updated in an outer loop as shown in Algorithm \ref{alg_AL}.

\subsection{Switching time optimization based on DDP}\label{sec_timeOpt}
While Algorithm \ref{alg_AL} finds the optimal control $\mathbf{U}^*$ for the OCP (\ref{eq_switchoptDT}) (equivalently (\ref{eq_optswitch})) for fixed $\tb$, the switching time optimization (STO) algorithm developed in this section updates $\tb$ toward an optimal value under a fixed control policy (\ref{eq_optdu}). This approach is different from \cite{xu2000dynamic}, where the control sequence $\mathbf{U}$ is fixed without feedback from the state.

The STO algorithm reformulates the OCP (\ref{eq_optswitch}) on fixed time intervals of length one, and augments the state vector with an extra state representing the time span of each mode. Denote $\z$ the time state, $\xbar$ the scaled system state, and $\X = [\xbar^T, \z^T]^T$ the augmented state. Then, Algorithm \ref{alg_AL} can be used to find $\V_{\xbar}$, $\V_{\xbar\xbar}$, $\V_{\z}$, $\V_{\z\z}$, and $\V_{\xbar\z}$ in the backward sweep. Following the convergence of Algorithm \ref{alg_AL}, the values of $\V_{\z}$ and $\V_{\z\z}$ are then used to update the switching times using Newton's method.

We first discuss the reformulation of the OCP (\ref{eq_optswitch}) on fixed time intervals and then the derivation of the STO under the new formulation.  Let $T_i=t_i-t_{i-1}$ and $\z = [T_1, \cdots, T_n]^T$. With the change of variable $\tau = {\frac{t-t_{i-1}}{T_i}+i-1}$, time-scaled dynamics are obtained as
\begin{equation}\label{eq_switchDyn_scale}
    \begin{cases}
    \dot{\xbar}(\tau) &= T_i\Bar{\f}_i(\xbar(\tau),\ub(\tau)),\ \tau\in[(i-1)^+, i^-] \\
    \xbar^+(\tau) &= \Pb_i(\xbar^-(\tau)), \ \tau\in[i^-, i^+]\\
    \dot{\z}(\tau) &= 0, \ \tau\in [0, n],
    \end{cases}.
\end{equation}
with the switching constraint
\begin{equation}\label{eq_constr_scale}
    g_{c_i}(\xbar(i^-)) = 0\,.
\end{equation}
The timing state $\z$ has the initial condition $\z(0) = \z_0$. The cost function in the OCP (\ref{eq_optswitch}) now becomes
\begin{equation}\label{eq_cost_scale}
\sum_{i=1}^{n} \left[\int_{(i-1)^+}^{i^-} T_il_i(\xbar(\tau),\ub(\tau)) {\rm d}\tau + \Phi_i(\xbar(i^-))\right].
\end{equation}

We can now apply Algorithm 1 to minimize (\ref{eq_cost_scale}) under the fixed initial condition $\z(0)$. Once Algorithm~\ref{alg_AL} converges, it implies that 1) the control sequence and trajectory are (locally) optimal and 2) the quadratic model of $V$ is a valid approximation of $V$. The gradient $\V_{\z}$ and Hessian $\V_{\z\z}$ are then obtained from the quadratic model. Since $\z$ only affects the dynamics via its initial condition, $\z(0)$ is updated using  
\begin{equation}\label{eq_time_update}
    \z_{\rm{new}}(0) = \z_{\rm{old}}(0) - \epsilon_z\V^{-1}_{\z\z}(0)\V_{\z}(0).
\end{equation}
where $0<\epsilon_z\leq1$ denotes the switching time line search parameter. Similar to DDP, we perform a backtracking line search to select $\epsilon_z$ and ensure cost reduction with (\ref{eq_time_update}).

\begin{algorithm}[t]
    \caption{STO algorithm}
    \label{alg_STO}
    \begin{algorithmic}[1] 
        \State \textbf{Given}
        \State Mode sequence $\mathcal{P}(n)$.
        \State Cost function (\ref{eq_cost_scale}), scaled dynamics (\ref{eq_switchDyn_scale}) and switching constraints (\ref{eq_constr_scale}).
        \State \textbf{Initialization}
        \State Initialize control sequence $\mathbf{U}$ and time state $\z_{0}$.
        \State \textbf{Execution}
        \State Execute Algorithm \ref{alg_AL} to obtain optimal control $\mathbf{U}^*$, feedback gain $\mathbf{K}$ in (\ref{eq_optdu}), $\V_{\z}$ and $\V_{\z\z}$.
        \State Line search using $\z(0) = \z(0) - \epsilon_z\V^{-1}_{\z\z}(0)\V_{\z(0)}$.
        \end{algorithmic}
\end{algorithm}

Algorithms \ref{alg_AL} and \ref{alg_STO} are combined to solve the OCP (\ref{eq_switchoptDT}) simultaneously for optimal $\mathbf{U}^*$ and $\tb^*$, as shown in Fig.~\ref{fig_HSDDP}, constituting HS-DDP. Note that the STO algorithm is executed after the AL algorithm converges, which implies that the feedforward term in equation~\eqref{eq_uff} becomes zero, and, thus, the control policy $\ub_k =\hat{\ub}_k+\mathbf{K}\delta\X_k$ is used in the line search for the timing variables and in the next forward sweep. The major difference between our method and the approach in \cite{xu2000dynamic} is the inclusion of this feedback term in the control law. The control policy used in this work allows (\ref{eq_time_update}) to make more aggressive updates, and consequently achieves faster convergence. The reason behind this is that any change in $\z(0)$ will create perturbations to the locally optimal trajectory. The effect of the change in $\z(0)$ on optimality is reduced by including the feedback term in control to account for perturbations. More details on this aspect are discussed in Sec.~\ref{sec_sim_STO}. 

\section{Results: Bounding with A 2D Quadruped }\label{sec_sim}

\subsection{Model and Simulation}

The developed HS-DDP algorithm is tested on a 2D model of the simulated MIT Mini Cheetah \cite{katz2019mini}, as in Fig.~\ref{fig_gait_cycle}. We consider two trajectory optimization tasks. The first task fixes the switching times and applies Algorithm \ref{alg_AL} on quadruped bounding for five gait cycles. We compare the results with those when AL + ReB is disabled and demonstrate satisfaction of constraints (\ref{subeq_switch}) - (\ref{subeq_friction}) within four AL iterations. The second task applies the HS-DDP to quadruped bounding for one gait cycle and demonstrates the efficiency of the STO.

\subsection{Five Gait Cycle Bounding with AL and ReB}
\begin{figure}[!b]
    \vspace{-8px}
    \centering
    \includegraphics[width=0.8\columnwidth]{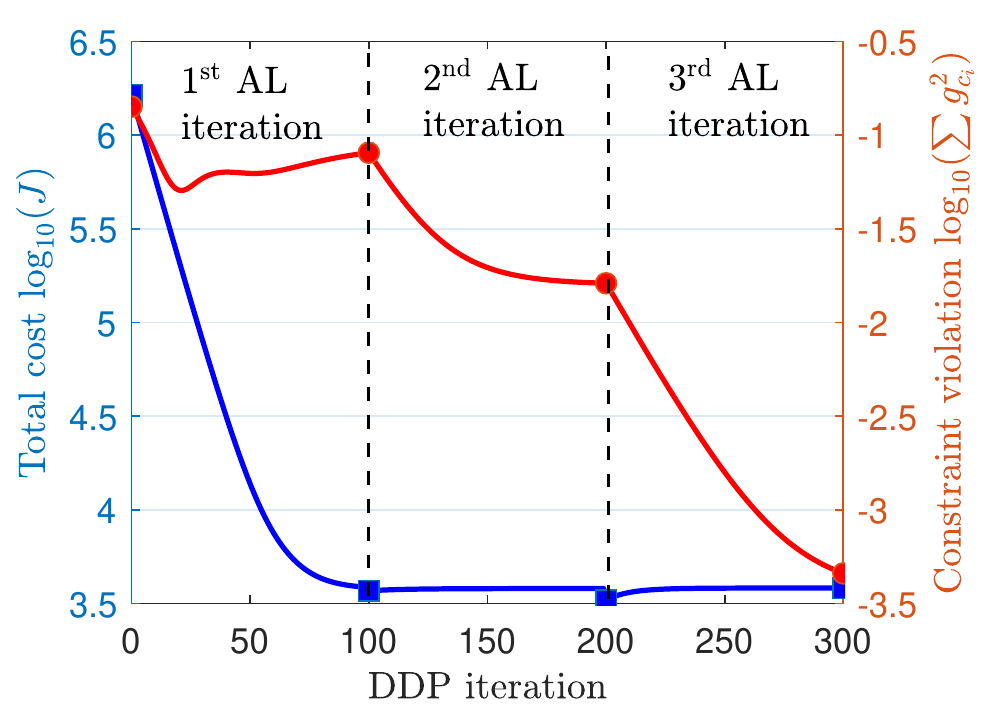}
    \vspace{-8px}
    \caption{Convergence of the total cost and constraint violation.}
    \label{fig_convergence}
\end{figure}

In this task, Algorithm \ref{alg_AL} is applied to 2D quadruped bounding for five gait cycles. The robot starts in the back-stance mode and is desired to run at an average forward speed of 1.0 m/s. A constant reference configuration is assigned to each mode, which mimics the robot's posture at the end of the mode and is selected heuristically. All desired joint velocities and the desired body vertical velocity are set to zero.

Quadratic running cost and terminal cost are used in (\ref{eq_optswitch}),
\begin{align}\label{eq_quadratic_cost}
l_i(\x,\ub) &= (\x-\x_{\text{ref},i})^T\mathbf{Q}_i(\x-\x_{\text{ref},i})+\ub^T\mathbf{R}_i\ub,\\
\Phi_i(\x(t_i)) &=  (\x(t_i)-\x_{\text{ref},i})^T\mathbf{Q}_{fi}(\x(t_i)-\x_{\text{ref},i}),
\end{align}
where $\mathbf{Q}_i$ and $\mathbf{R}_i$ are weighting matrices for state deviation and energy consumption in running cost, respectively, and $\mathbf{Q}_{fi}$ is the weighting matrix for the terminal cost (of the $i^{\text{th}}$ mode). In this simulation, we have zero penalty on forward position, and relatively larger penalty on forward speed, body height, and joint velocities than the other states. The integration time step $h=1$ ms is used, and the switching times are selected such that the flight mode (and the front-stance mode) runs for 72 ms and the back stance mode runs for 80 ms. The initial guess for Algorithm \ref{alg_AL} is given by a heuristic controller, which implements the PD control in flight mode such that a predefined joint configuration is maintained. In stance mode, the heuristic controller constructs stance leg forces following a SLIP model and converts the Ground Reaction Force (GRF) thus obtained to joint torques. The AL and ReB parameters are initialized as $\sigma = 5$, $\lambda_i = 0$, $\beta = 8$, and $\sigma = 0.5$, and the convergence threshold is set to $\epsilon_{AL} = 10^{-4}$.

\subsection{AL and ReB Simulation Results}
\label{subsec_AL_and_ReB}
When AL is active and ReB is disabled, it takes three AL iterations for the constraint violation to decrease within $\epsilon_{AL}$. The convergence of the total cost (excluding the penalty term and the Lagrangian term) and switching constraint violation are shown in Fig.~\ref{fig_convergence}. The blue square markers and the red circle markers indicate the beginning of the corresponding AL iteration. Figure \ref{fig_convergence} demonstrates that at least one of the total cost and the constraint violation is reduced at every DDP iteration. Further, the algorithm spends more effort in minimizing the total cost at the beginning and switches to the constraint violation after the total cost is converged.
\begin{figure*}[!t]
    \centering
    \includegraphics[width=0.7\textwidth]{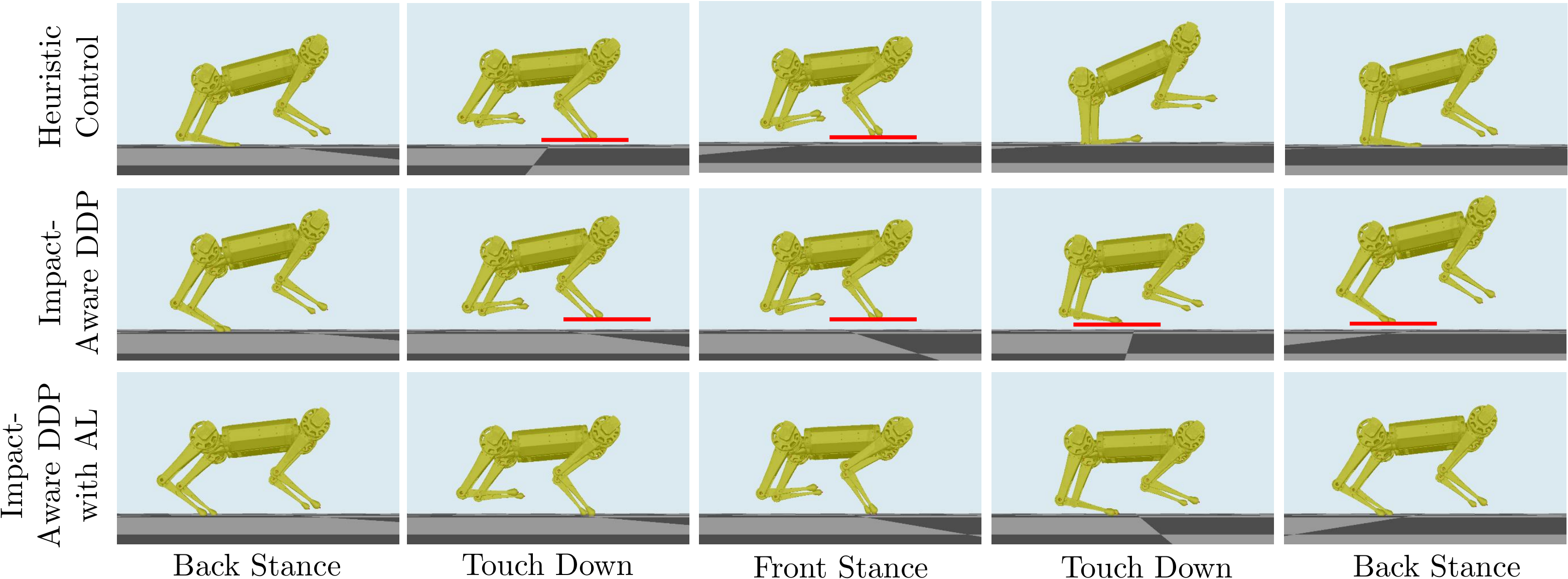}
    \vspace{-8px}
    \caption{Sequential snapshots of the generated bounding motion for Mini Cheetah. Top: Motion generated by the heuristic controller that is used to warm start AL. Middle: Motion generated by DDP (without AL) ignoring switching constraints. Bottom: Motion generated with AL enforcing switching constraints. The first two methods incorrectly regard the red lines as the ground, and thus, dynamics are reset on this `virtual ground'.}
    \label{fig_motion}
\end{figure*}

Figure \ref{fig_motion} compares the bounding gaits that are generated by three methods: 1) A heuristic controller that is used to warm start the optimization, 2) DDP (with impact-aware value function update) that ignores switching constraints, and 3) AL that enforces switching constraints. It demonstrates that the developed AL algorithm achieves the desired performance. Though the motion generated by DDP is more smooth and realistic compared to the heuristic controller, the robot still violates the switching constraints, and the error accumulates over time. This behavior is because the first two methods do not enforce switching constraints, and thus, the robot does not correctly recognize the ground, but the simulator still resets the dynamics.

\begin{figure}[!b]
    \centering
    \vspace{-8px}
    \includegraphics[width=\columnwidth]{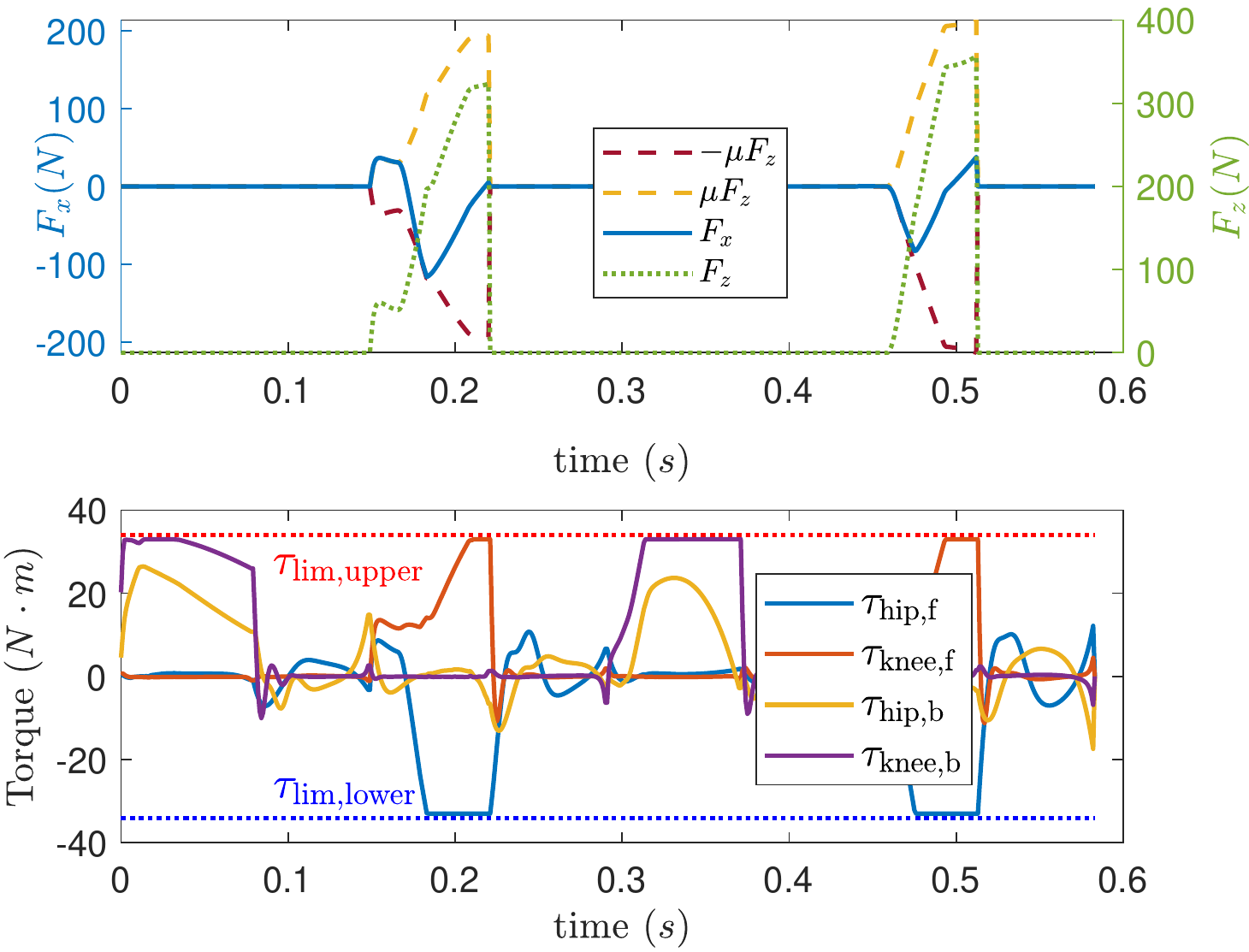}
    \caption{ GRF and joint toques for 2D Mini Cheetah bounding. Top: Normal and tangential GRF for the front leg. Bottom: Joint torques. With AL and ReB, the non-negativity of normal GRF, friction, and torque limit constraints are satisfied in four AL iterations.}
    \label{fig_ineq_constr}
\end{figure}

Figure \ref{fig_ineq_constr} depicts the normal and tangential GRF for the front leg (top), and the torques for every joint (bottom) when the ReB is activated. The algorithm terminates at four AL iterations. It shows that the normal GRF is always non-negative and that the friction and joint torques are confined within their boundaries, demonstrating effectiveness of the ReB method. Similar results are observed for the back leg.

\subsection{One-Gait Bounding with Time Optimized}\label{sec_sim_STO}
In this task, HS-DDP is applied to the generation of one bounding gait for the Mini Cheetah. Different from the previous task, where only the control is optimized, switching times are also optimized in this task. Only one gait cycle is studied here in the sense that, in many situations, the switching times found for one gait cycle can be extended to the succeeding gait cycles. The cost function, initial control sequence,  initial switching times,  AL parameters, and terminating conditions all remain the same in this task as in the previous one. 

\subsection{HS-DDP Simulation Results}
\begin{figure}[!b]
    \centering
    \includegraphics[width=0.82\columnwidth]{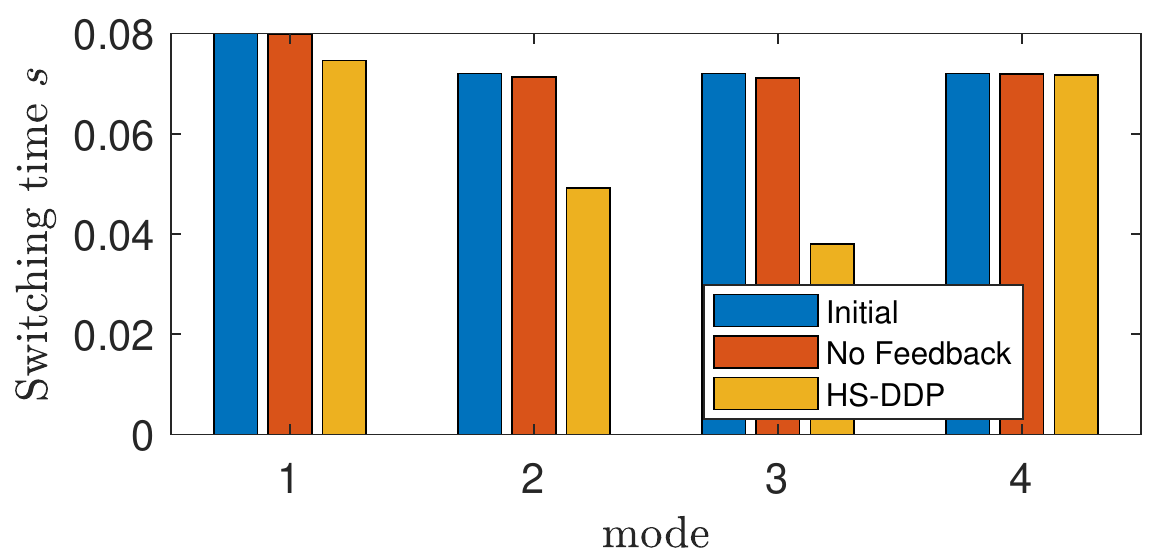}
    \vspace{-8px}
    \caption{Time spent in each mode for the one-gait-cycle bounding task.}
    \label{fig_time}
\end{figure}

The optimal switching times obtained via the STO algorithm in HS-DDP are shown in Fig.~\ref{fig_time}. The algorithm reduces the time of the first flight mode and the front-stance mode. 
Figure \ref{fig_time} also compares the switching times obtained via the STO algorithm with the algorithm proposed in \cite{xu2000dynamic} where the feedback control is not used. Both algorithms are terminated at the $30^{\rm{th}}$ iteration. With HS-DDP, the overall time spent on the entire motion is 0.2335 s, a $21.1\%$ reduction of the initial overall time, whereas only a $6.3 \%$ reduction is observed with the algorithm in \cite{xu2000dynamic}, showing that the HS-DDP is more efficient in the sense of taking larger steps.

\begin{figure}[!t]
    \centering
    \includegraphics[width = 0.9\columnwidth]{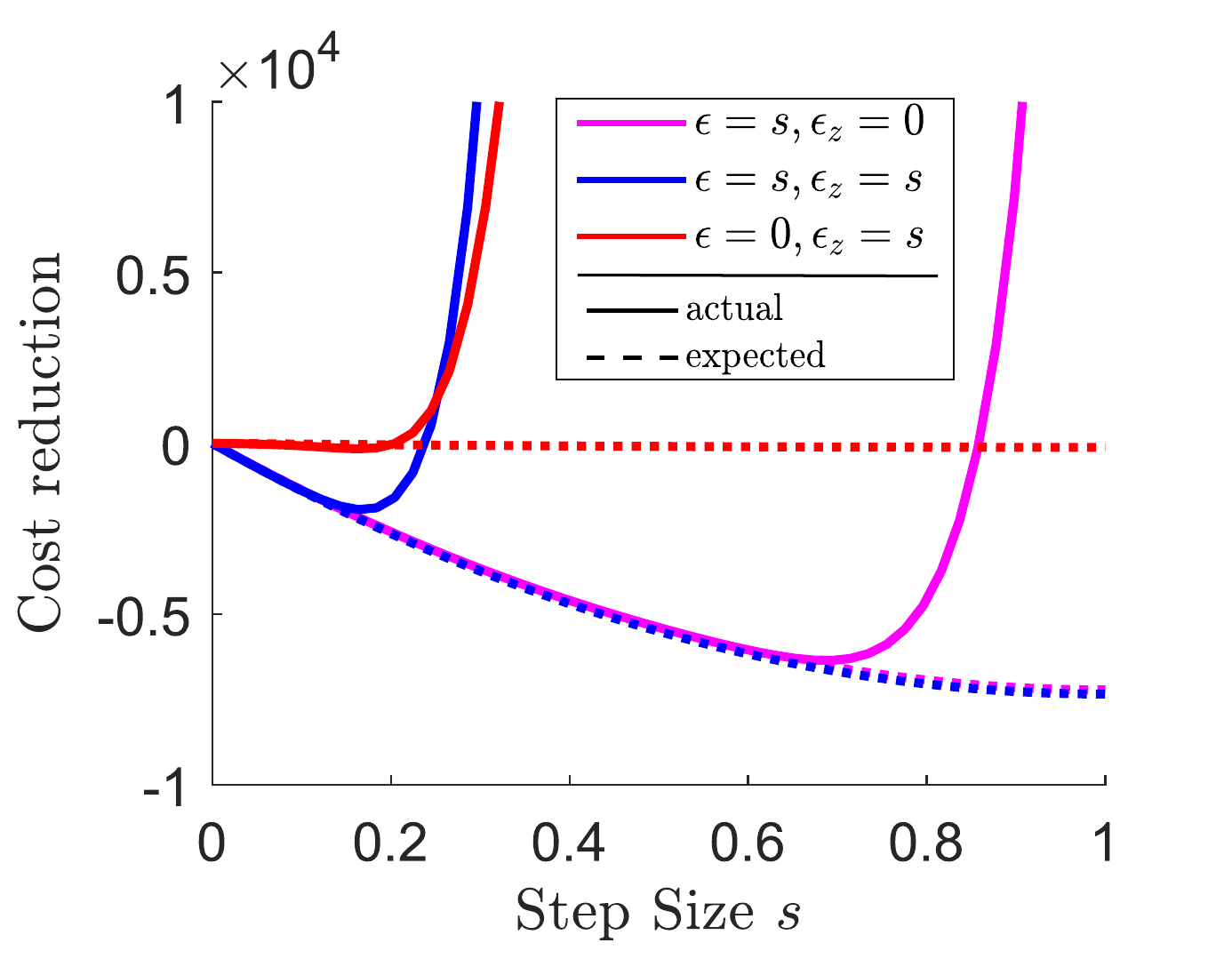}
    \vspace{-8px}
    \caption{Change in cost with respect to step size. Solid lines: actual cost reduction. Dashed lines: predicted cost reduction. Red: $\epsilon=0$, $\epsilon_z = s$. Pink: $\epsilon=s$, $\epsilon_z = 0$. Blue: $\epsilon = \epsilon_z = s$.}
    \label{fig_cost_reduction}
    \vspace{-10px}
\end{figure}

Figure \ref{fig_cost_reduction} explains why the two-level optimization strategy is adopted in HS-DDP. With the scaled optimization structure (\ref{eq_switchDyn_scale}), (\ref{eq_constr_scale}), and (\ref{eq_cost_scale}), it is reasonable to update the control using \eqref{eq_uff} and the switching times using (\ref{eq_time_update}) simultaneously since the gradient and Hessian information are all available in the backward sweep of DDP. If the actual cost reduction is less than zero and is close to the predicted cost reduction, then the quadratic model of the value function is considered valid. The quadratic approximation, however, is more sensitive to the switching time line search parameter $\epsilon_z$ than the control line search parameter $\epsilon$, as shown in Fig.~\ref{fig_cost_reduction}. This figure indicates that $\epsilon$ has to be as small as $\epsilon_z$ if \eqref{eq_uff} and (\ref{eq_time_update}) are executed simultaneously, at the price of much less cost reduction per iteration, thus decreasing the convergence rate. Although this behavior is not observed for all iterations, it can significantly slow down the optimization if a small step size is continuously used for multiple iterations.

\section{Conclusions and Future Works}\label{sec_conclusion}
The proposed HS-DDP framework combines three algorithmic advances to DDP for legged locomotion. It addresses the discontinuity at impacts by incorporating an impact-aware value function update in the backward sweep. By combing AL and DDP, HS-DDP reduces either the total cost or the constraint violation in every iteration, enforcing the switching constraint as the algorithm proceeds. Further, with the developed STO algorithm, HS-DDP can efficiently find the optimal switching times alongside the optimal control. A ReB method is combined with HS-DDP to manage the inequality constraints. The five-gait-cycle bounding example shows the promise of HS-DDP in rapidly satisfying the switching constraint in just a few iterations, and demonstrates the effectiveness of ReB for enforcing inequality constraints. The one-gait-cycle bounding example compares the developed STO algorithm to the previous solutions, demonstrating that our method is more efficient due to the inclusion of the feedback control in the forward sweep.

Though forward Euler integration is used in this work for dynamics simulation, the developed HS-DDP is independent of the integration scheme. Implicit or higher-order methods can be used if the computation time is not the primary concern. The current implementation of HS-DDP is MATLAB based, and so future work will benchmark its computational performance with C++ and realize the developed algorithm in experiments for real-time control with the Mini Cheetah. We are also interested in comparing ReB and AL in terms of their abilities for inequality constraint management.

\ifarxiv
\bibliography{ResearchGate.bbl}
\else
\bibliographystyle{IEEEtran}
\bibliography{paperbib}
\fi

\end{document}